\documentclass[10pt,twocolumn,letterpaper]{article}

\usepackage{iccv}
\usepackage{times}
\usepackage{epsfig}
\usepackage{graphicx}
\usepackage{amsmath}
\usepackage{amssymb}
\usepackage[dvipsnames]{xcolor}
\usepackage{booktabs}
\usepackage{cuted}
\usepackage{capt-of}
\usepackage{soul}
\usepackage[numbers,sort]{natbib}

\newcommand{\methodNameShort}{SAT}

\usepackage{subfig}

\usepackage[pagebackref=true,breaklinks=true,colorlinks,bookmarks=false]{hyperref}

\iccvfinalcopy %

\date{\includegraphics[height=2in]{example-image}}

\begin{document}

\title{Aligning Subtitles in Sign Language Videos}

\author{Hannah Bull$^{1}$\thanks{Equal contribution} \quad Triantafyllos Afouras$^{2*}$ \quad G\"ul Varol$^{2,3}$  \\ Samuel Albanie$^{2}$  \quad Liliane Momeni$^{2}$ \quad Andrew Zisserman$^2$\\
$^{1}$ LISN, Univ Paris-Saclay, CNRS, France \\
$^{2}$ Visual Geometry Group, University of Oxford, UK \\
$^{3}$ LIGM, \'Ecole des Ponts, Univ Gustave Eiffel, CNRS, France \\
{\tt\small hannah.bull@lisn.upsaclay.fr; \{afourast,gul,albanie,liliane,az\}@robots.ox.ac.uk}\\
{\tt\small \url{https://www.robots.ox.ac.uk/~vgg/research/bslalign/} }
}

\maketitle

\begin{strip}\centering
\vspace{-15mm}
\includegraphics[width=\textwidth]{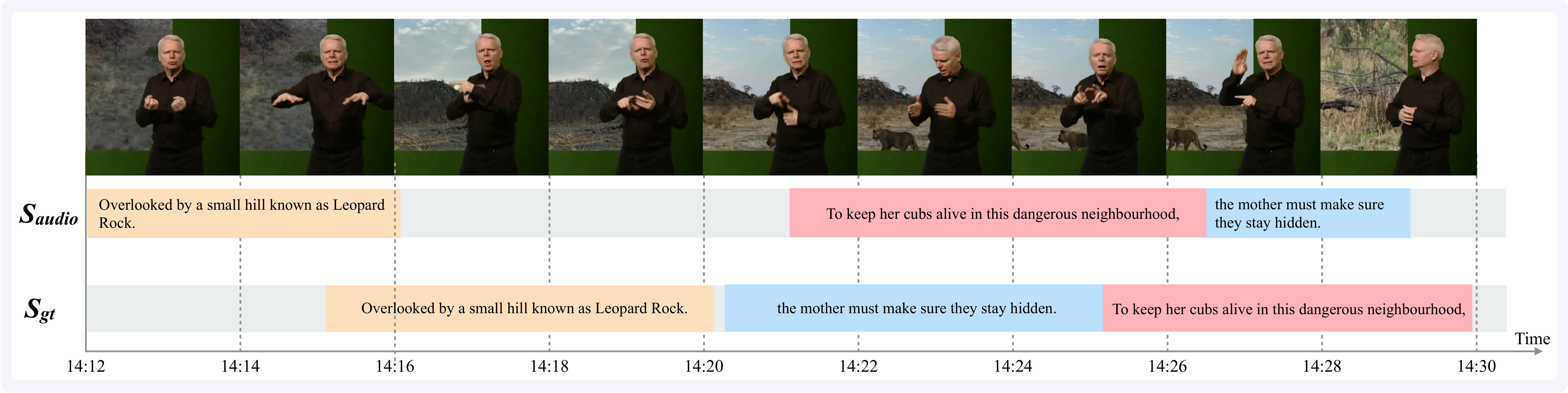}
\mbox{}\vspace{-.7cm}\\
\captionof{figure}{\textbf{Subtitle alignment}: We study the task of aligning subtitles to continuous signing in sign language interpreted TV broadcast data. The subtitles in such settings usually correspond to and are aligned with the audio content (top: audio subtitles, {\bf $S_{audio}$}) 
but are unaligned with the accompanying signing (bottom: Ground Truth annotation of the signing corresponding to the subtitle,  {\bf $S_{gt}$}). 
This is a \textit{very challenging} task as (i)~the \textit{order} of subtitles varies between spoken and sign languages, (ii)~the \textit{duration} of a subtitle differs considerably between signing and speech, and (iii)~the signing corresponds to a \textit{translation} of the speech as opposed to a transcription. 
\label{fig:teaser}}
\end{strip}

\begin{abstract}

The goal of this work is to temporally
align asynchronous subtitles in sign language videos.
In particular, we focus on sign-language
interpreted TV broadcast data comprising
(i) a video of continuous signing, and (ii) subtitles
corresponding to the audio content.
Previous work exploiting such weakly-aligned
data only considered finding keyword-sign correspondences,
whereas we aim to localise a complete
subtitle text in continuous signing.
We propose a Transformer architecture
tailored for this task, which we train on 
manually annotated alignments covering over 15K subtitles
that span 17.7 hours of video.
We use BERT subtitle embeddings
and CNN video representations learned for sign recognition
to encode the two signals, which interact through
a series of attention layers. %
Our model outputs
frame-level predictions, i.e.,
for each video frame, whether it belongs to the queried subtitle or not.
Through 
extensive evaluations, we show
substantial improvements over existing alignment baselines that do not make use of  subtitle text embeddings for learning. Our automatic alignment
model opens up possibilities for advancing
machine translation of sign languages via
providing continuously synchronized video-text data.

\end{abstract}
\section{Introduction}
\label{sec:intro}

Sign languages constitute a key form of communication for Deaf 
communities~\cite{sutton-spence_woll_1999}.
Our goal in this paper is to temporally localise subtitles in continuous 
signing video. Automatic alignment of subtitle text to signing content has 
great potential for a wide range of applications including
assistive tools for education and translation,
indexing of sign language video corpora,
efficient subtitling technology for signing vloggers\footnote{Unlike 
spoken vlogs that benefit from automatic closed captioning on sites such as 
YouTube, signing vlog creators %
who wish to provide written subtitles %
must both 
translate \textit{and} align their subtitles manually.}, and
~automatic construction
of large-scale sign language datasets that support computer vision and linguistic research.

Despite recent advances in computer vision, machine translation between 
continuous signing and written language remains largely 
unsolved~\cite{bragg2019sign}. 
Recent works~\cite{Camgoz_2018_CVPR,Camgoz_2020_CVPR} %
have shown promising translation results, but to date these have been achieved only %
in 
\textit{constrained} settings where continuous signing is \textit{manually 
pre-segmented} into clips, with each clip associated to a written sentence from 
a \textit{limited vocabulary}.
Two key bottlenecks for scaling up translation 
to continuous signing depicting unconstrained vocabularies are (i)~the 
segmentation of signing into sentence-like units, and (ii)~the availability of 
large-scale sign language training data.

Manual alignment of subtitles to sign language video is tedious -- an expert fluent in sign language takes approximately 10-15 hours to align subtitles to 1 hour of continuous sign language video. In this work, we focus on the task of aligning a particular known subtitle 
within a given temporal signing window. We explore this task in the context of 
sign language interpreted TV broadcast footage -- a readily available and 
large-scale source of data -- where the subtitles are synchronised with the 
audio, but the corresponding sign language translations are largely unaligned 
due to differences between spoken and sign languages as well as lags from the 
live interpretation. 

Subtitle alignment to continuous signing %
remains a \textit{very challenging} task. First, sign languages have grammatical structures 
that vary considerably from those of spoken languages~\cite{sutton-spence_woll_1999}, and as 
a result the \textit{ordering} of words within a subtitle as well as the subtitles 
themselves is often not maintained in the signing (see Fig.~\ref{fig:teaser}). Second, the 
\textit{duration} of a subtitle varies considerably between signing and speech due to 
differences in speed and grammar. Third, the signing corresponds to a \textit{translation} 
of the speech that appears in the subtitles as opposed to a transcription: there 
is no direct one-to-one mapping between subtitle words and signs produced by interpreters,
and entire subtitles may not be signed.

Previous work exploiting such weakly-aligned data has mainly focused on finding sparse 
correspondences between keywords in the subtitle and individual 
signs~\cite{Albanie20,Momeni20b,Varol21}, as opposed to localising the start and end times 
of a complete subtitle text in continuous signing. Though, as we show, localising isolated 
signs identified by keyword spotting nevertheless forms a useful pretraining task for full 
subtitle alignment. Most closely related to our work, Bull~et~al.~\cite{bull2020} consider 
the task of segmenting a continuous signing video into subtitle units purely based on body 
keypoints. In fact, similarly to speech which can be segmented based on prosodic cues such 
as pauses, sign sentence boundaries can\textit{ to an extent} be detected through visual 
cues such as lowering the hands, head movement, pauses, and facial expressions \cite{Fenlon2010SeeingSB}.
However, as shown in our evaluations in Sec.~\ref{sec:experiments}, such approaches based on 
prosody-only perform poorly in our setting, where subtitles do not necessarily correspond to 
complete sign sentences with clear visual boundaries.

In this paper, we instead propose to use \textit{the subtitle text as an additional signal} 
for better alignment. We make the following three contributions:
(1) we show that encoding the subtitle text
as input to the alignment model significantly
improves the temporal localisation quality
as opposed to only relying on visual cues
to segment continuous sign language videos into subtitle units;
(2) we design a novel formulation
for the subtitle alignment task based on
Transformers; %
and (3) we present a comprehensive study
ablating our design choices and provide
promising results for this new task
when evaluating on unseen signers and content.

\section{Related Work}
\label{sec:relatedwork}
For a recent comprehensive survey about
sign language recognition and translation,
see \cite{koller2020quantitative}.
Here, we review relevant works on
temporal localisation at the levels of individual signs and sequences,
in addition to
more general temporal alignment methods from the literature.

\noindent\textbf{Temporal localisation of individual signs.} A rich body of work has 
considered the task of localising sparse sign instances in continuous signing, often 
referred to as ``sign spotting''. Early efforts using signing gloves~\cite{liang1998real} 
were followed by methods employing hand-crafted visual features to represent the hands, face 
and motion that were integrated with CRFs~\cite{yang2006detecting,yang2008sign}, 
HMMs~\cite{Santemiz2009AutomaticSS} and HSP Trees~\cite{ong2014sign}.
Several studies have sought to employ subtitles as weak supervision for learning to localise 
and classify signs, using apriori mining~\cite{cooper2009learning} and multiple-instance 
learning~\cite{Buehler09,Buehler10,Pfister13}.
More recent work has leveraged cues such as mouthings~\cite{Albanie20} and visual 
dictionaries~\cite{Momeni20b} and by making use of deep neural network features with sliding 
window classifiers~\cite{li2020transferring} and attention learned via a proxy translation 
task~\cite{Varol21}. In deviation from these works, our objective is to localise complete 
subtitle units, rather than individual signs.

\noindent\textbf{Temporal localisation of sign sequences.} The alignment of subtitles to 
continuous signing was considered in creative early work by combining cues from multiple 
sparse correspondences~\cite{farhadi2006aligning}, but under the assumption that ordering of 
words in subtitles are preserved in the signing (which does not hold in our problem setting).
 \begin{figure*}
    \centering
    \includegraphics[width=\textwidth]{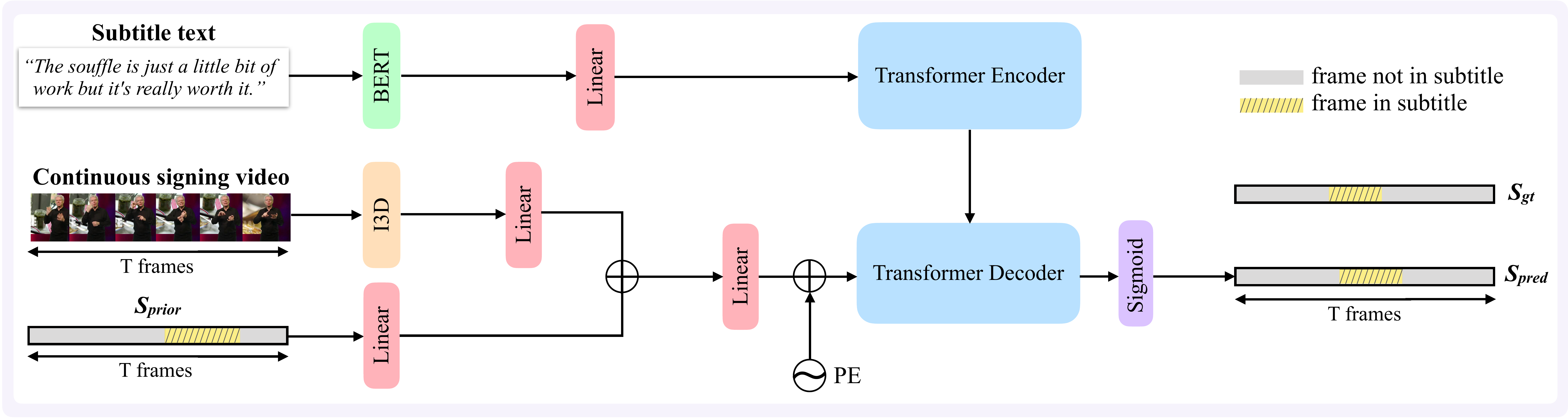}
    \caption{
        \textbf{SAT model overview: } We input to our model (i) token embeddings of the subtitle text we wish to align, (ii) a sequence of video features extracted from a continuous sign language video segment and (iii) the shifted temporal boundaries of the audio-aligned subtitle, S$_{prior}$. Using these inputs, the model outputs a vector of values between 0 and 1 of length $T$. Its first and last values above a threshold $\tau$ delimit the predicted temporal boundaries for the query subtitle. The location of the subtitle with respect to the window is represented in dashed yellow.
    }
    \label{fig:model}
\end{figure*}
Other sequence-level sign language temporal localisation tasks that have received attention 
in the literature include category-agnostic sign 
segmentation~\cite{farag2019learning,Renz21a}, active signer 
detection~\cite{Cherniavsky2008,Borg2019SignLD,moryossef2020real,Shipman2017SpeedAccuracyTF} 
and diarisation~\cite{gebrekidan2013automatic,gebre2014motion,albanie21_seehear}---each 
considers a temporal granularity that differs from subtitle units.
Most closely related to our work, Bull~et~al.~\cite{bull2020} employ a keypoint-based model to 
segment continuous signing into sentence-like units without knowledge of the written 
subtitles during inference. Our approach relaxes this assumption and considers instead the 
practical scenario
in which we assume access to the written subtitle to be aligned. We compare our approach with theirs in Sec.~\ref{sec:experiments}.

\noindent\textbf{Continuous sign language recognition.}
Hybrid models coupling CNNs with HMMs~\cite{koller2016deep,koller2017re}, attention 
mechanisms~\cite{huang2018video} and CTC losses~\cite{camgoz2017subunets,cheng2020fully} 
have been studied for continuous sign language recognition, with recent extensions to 
sequence-to-sequence models~\cite{Camgoz_2018_CVPR} and 
Transformers~\cite{Camgoz_2020_CVPR,Li2020TSPNetHF} to tackle the task of sign language 
translation. These models produce either implicit or explicit alignments over a signing 
sequence corresponding to a sentence. However, these approaches have only been demonstrated 
to work on \textit{pre-segmented} sentences of signing~\cite{Camgoz_2018_CVPR}.

\noindent\textbf{Aligning bodies of text to video.}
The Dynamic Time Warping (DTW) algorithm~\cite{Myers1981ACS} has been applied to the problem of 
aligning sequences of movies to 
transcripts~\cite{Everingham2006HelloMN,Sankar2009SubtitlefreeMT} and plots 
synopses~\cite{Tapaswi2014StorybasedVR} using cues such as character recognition and 
subtitle content. It has also been successfully applied to the problem of aligning generic 
text descriptions against untrimmed video~\cite{Bojanowski2015WeaklySupervisedAO}. While 
effective, these methods require the preservation of sequence ordering across modalities, 
which does not hold in our problem setting. We nevertheless show in Sec.~\ref{sec:method} 
how DTW can be used as a secondary stage of processing that resolves conflicting local 
alignments on the re-ordered subtitle prediction timings via a global objective. The fixed 
ordering assumption is relaxed by the work of~\cite{Tapaswi2015Book2MovieAV}, which aligns 
book chapters to video scenes. Their approach, however, which works through matching sparse 
character identifications against specific shots, is not applicable in our setting where 
shot boundaries do not provide a natural segmentation of the signing content.

\noindent\textbf{Natural language grounding in videos.} Our work is also related to the task 
of natural language grounding, which aims to locate a temporal segment within an untrimmed 
video sequence corresponding to a given natural language query. Existing methods have 
considered two-stage \textit{propose and rank} 
approaches~\cite{Hendricks2017LocalizingMI,gao2017tall,Liu2018AttentiveMR,Xu2019MultilevelLA}, 
iterative grounding agents trained with reinforcement 
learning~\cite{He2019ReadWA,Wang2019LanguageDrivenTA} and single-stage regression 
models~\cite{Yuan2019ToFW,Ghosh2019ExCLEC,Chen2018TemporallyGN,Zeng2020DenseRN}.
Our proposed subtitle alignment task differs from natural language grounding in three ways: 
(i) The signing content is more \textit{fine-grained}---the visual appearance of a signing 
sequence remains very similar across frames, necessitating nuanced %
recognition of body dynamics;
(ii) Differently from language grounding, each subtitle to be aligned comes with its own 
reference location, providing an instance-specific prior over the start time and duration.
As we show in Sec.~\ref{sec:experiments}, our effective use of this reference is important to 
achieving good performance, and our model is specifically designed to take advantage of this 
cue; (iii) Subtitles occupy mutually exclusive temporal regions, a property that we further 
exploit to improve alignment quality, but that does not hold in general for natural language grounding.

\section{Method}
\label{sec:method}
In this section, we describe our Transformer-based subtitle alignment  model operating on a 
single subtitle and a short video segment (Sec.~\ref{subsec:sat}), our pretraining on sparse 
sign spottings (Sec.~\ref{subsec:wordpretraining}), and our final step that globally adjusts 
multiple subtitles in a long video using DTW (Sec.~\ref{subsec:dtw}).

\noindent\textbf{Problem formulation.}
As inputs to the model, we provide
(i)~token embeddings of the subtitle text we wish to align to signing,
(ii)~a sequence of video features extracted from a continuous sign language video segment,
as well as 
(iii) prior estimates of the temporal boundaries for the given query, which we refer to as S$_{prior}$.
The latter is provided as an approximate location and duration cue of 
the signing-aligned subtitle.
Using these inputs, we predict a binary vector of the same length as the video features,
where a consecutive sequence of 1s denotes the temporal location of the subtitle.

\subsection{Subtitle Aligner Transformer}
\label{subsec:sat}
The core of our model is a Transformer~\cite{vaswani2017attention}, as shown in 
Fig.~\ref{fig:model}, which we refer to as Subtitle Aligner Transformer (SAT).
In contrast to the common approach of feeding video
frames as input to the encoder \cite{desai2021virtex,carion2020end}, we input the video
frames to the \textit{decoder} side in order for the model to learn the association between 
the frame-level features and the output vector of the same duration.
We first describe the structure of the Transformer, and then the text and video feature extraction.
Additional implementation details are provided in Sec.~\ref{app:sec:implementation} of the appendix.

\noindent\textbf{Encoder.}
The input to the encoder is a sequence of text embeddings corresponding to the subtitle we wish to align.
Positional encodings are not used on the encoder side of
the Transformer since the text embeddings (see below) already contain
positional information.
The encoder is a stack of Transformer layers, each containing a multi-head attention mechanism followed by a feedforward network and embedding dimensionalities of size $d_{model}$.

\noindent\textbf{Decoder.}
The decoder is a stack of Transformer layers that attend on the encoded sequence.\footnote{Note: There is no auto-regression.} 
The input to the decoder consists of a sequence of video features encoding the visual signing 
information from the video, as well as a binary vector representing a prior estimation of the location 
of the signing-aligned subtitle (S$_{prior}$).
Positional encodings are added to the decoder input in order for the model to exploit the temporal ordering of the signing. 
The final layer of the model is a linear layer with a sigmoid activation which outputs $T$
predictions in the range $[0,1]$ one for each video frame.
Values of this output vector, S$_{pred}$, that are above a threshold $\tau$ correspond
to the predicted temporal location of the queried subtitle text.

\noindent\textbf{Text features.}
Each subtitle is encoded using a BERT~\cite{devlin2019bert} model pretrained on a large text corpus with a masked language modelling task, to produce a sequence of
768-dimensional vectors, one for each %
token in the sentence. 
To match the input dimension of the encoder Transformer, these embeddings are first linearly projected to $d_{model}$.

\noindent\textbf{Video features.}
The visual features are 1024-dimensional embeddings extracted from the I3D~\cite{carreira2017quo} sign 
classification model made publicly available by the authors of~\cite{Varol21}.
The features are pre-extracted over sign language video segments. 
A visual feature sequence of length $T$ is used as input to the model.

\noindent\textbf{Prior position encoding.}
Besides the video features, the input to the decoder also includes a subtitle timing estimate as a prior position and duration cue.
This prior estimate is encoded as a binary vector of length $T$, where 1 indicates that the associated video frame is within the temporal boundaries of the subtitle, %
and 0 otherwise.
The video and prior inputs are fused via concatenation before being passed as input to the decoder.
Before the concatenation both inputs are linearly projected to the same dimension. 
The fusion output is finally projected to $d_{model}$ in order to be input to the Transformer decoder.

\noindent\textbf{Training objective.}
The model is trained with a binary cross entropy loss between the 
predicted vector and the ground truth S$_{gt}$ of the signing-aligned subtitle within the video segment:
$$
\mathcal{L}
=
-
\frac{1}{T}
\sum_{t=1}^{T}
S_{gt}^t \log S_{pred}^t + (1-S^t_{gt}) \log (1-S_{pred}^t).
$$

\subsection{Word pretraining with individual sign locations}
\label{subsec:wordpretraining}

SAT is designed for alignment of subtitles to video signing streams. 
However, the same architecture can be used without any alterations
to align smaller text units, e.g.\ single words. 
Given that we have access to sparse sign annotations from mouthings \cite{Albanie20} and dictionary 
exemplars~\cite{Momeni20b}, we can use these to initialise the model weights and
incorporate this 
knowledge via a potentially easier single-sign spotting task. We obtain timings of the
sparse word-level 
annotations and assume a fixed single-second width as the precise sign boundaries are
not available.
The model is then trained to spot the single sign 
occurrence within a video window of size $T$. In our experiments, we demonstrate the advantages of such a pretraining strategy.

\subsection{Global alignment with DTW}
\label{subsec:dtw}

Our model does not take into account global information from the length of the video (e.g.\ 1-hour),
rather it looks for signing associated to a given subtitle within a short temporal window $T$ (e.g.\ 20-seconds). Hence, there may be overlaps between predictions for different subtitles; we resolve these overlap conflicts using DTW~\cite{Myers1981ACS}. 
We find an order-preserving global alignment from all elements of a sequence of video frames to all elements of sequence of subtitles, maximising the sum of sigmoid outputs of our model in our cost function for each subtitle query. 

As DTW aligns all frames in a video sequence to subtitles, we select all frames of the signing video which are likely to be associated with subtitle queries. Specifically, we select all frames associated to an output score over $\tau_{dtw}$. In the case where our model outputs only values below $\tau_{dtw}$ for a particular subtitle, we instead select all frames within the prior location S$_{prior}$. 

We order the subtitles by the mid-point of their predicted temporal location.
This allows the predicted subtitles to follow a different order to the original subtitles, because the order of phrases in the sign language interpretation does not necessarily follow the order of phrases of the written English subtitles (see  Sec.~\ref{app:sec:experiments} of the appendix for further details). 

We construct a cost matrix of dimension (i) the number of frames by (ii) the number of subtitles, and with entries of 
$1-p_{ij}$, where $p_{ij}$ is the sigmoid output corresponding to frame $i$ with subtitle $j$ as the encoder input. We apply the 
DTW algorithm to this cost matrix of aligning video frames to subtitles. This maximises the overall sum of the 
sigmoid outputs of the model under the ordering and allocation constraints of DTW. 

If not otherwise mentioned, our full SAT model uses DTW postprocessing.

\section{Experiments}
\label{sec:experiments}

In this section, we first give implementation details (Sec.~\ref{subsec:impl_details}) and describe the datasets and evaluation metrics
used in this work (Sec.~\ref{subsec:data}).
We then compare the results of the proposed \methodNameShort{} model against strong 
baselines (Sec.~\ref{subsec:baselines})
and present a series of ablation studies (Sec.~\ref{subsec:ablations}).
Next, we demonstrate the performance of our model
on an additional dataset (Sec.~\ref{subsec:bslcp}).
Finally, we provide qualitative results and discuss limitations (Sec.~\ref{subsec:qualitative}).

\subsection{Implementation details}
\label{subsec:impl_details}
\noindent\textbf{Architecture.} For both the encoder and the decoder we use 2 identical Transformer layers with 2 heads and size $d_{model}=512$ each.

\noindent\textbf{Backbone pretraining.}
The I3D model is pretrained to perform 1064-way classification across the sign spotting 
instances with mouthings~\cite{Albanie20} and dictionary exemplars~\cite{Momeni20b} (further details 
can be found in~\cite{Varol21}).
The model is then frozen and used to densely pre-extract visual features with stride 1 over the clips of the datasets.

\noindent\textbf{Prior input selection.}
As the prior estimate input S$_{prior}$ we use the temporal location of the audio-aligned subtitle S$_{audio}$
shifted by +3.2 seconds. This value, which we denote with S$_{audio}^{\text{+}}$, corresponds to the average temporal shift between the audio-aligned subtitles S$_{audio}$ and the ground truth 
subtitles S$_{gt}$ in our training data (see Fig.~\ref{fig:distribution_start}).

\noindent\textbf{Search windows.}
During training, we randomly select a search window of 20 seconds around the location of the ground truth subtitle S$_{gt}$, select the densely extracted video features for this window, and temporally subsample them by a factor of 4.
All videos are sampled at $25$ FPS, therefore this results in $T=125$ frames. 
During testing, we select a search window of the same length centered around the shifted subtitle location S$_{audio}^{\text{+}}$.

\noindent\textbf{Text augmentation.}
During training, we augment the text query inputs randomly to reduce overfitting: For 50\% of the samples, we shuffle the word order and add or delete up to two words.

\noindent\textbf{Hyper-parameters.} We set thresholds $\tau$ to $0.5$, $\tau_{dtw}$ to $0.4$. 
Further details are provided in appendix Sec.~\ref{app:sec:implementation}.

\subsection{Data and evaluation metrics}
\label{subsec:data}

\begin{figure}
    \centering
    \subfloat[]{\label{fig:distribution_start}\includegraphics[width=0.51\linewidth]{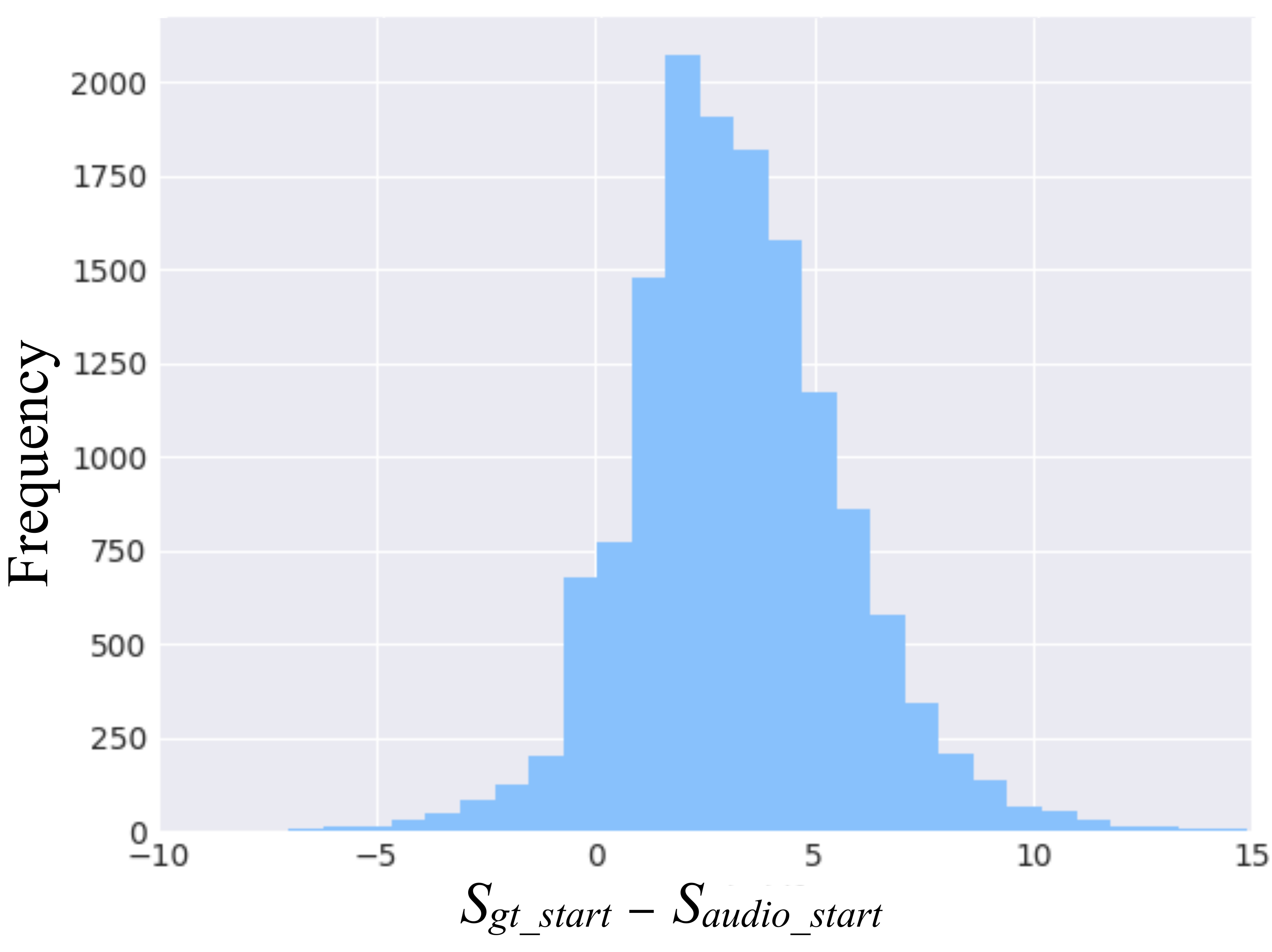}}\hfill
    \subfloat[]{\label{fig:distribution_dur}\includegraphics[width=0.489\linewidth]{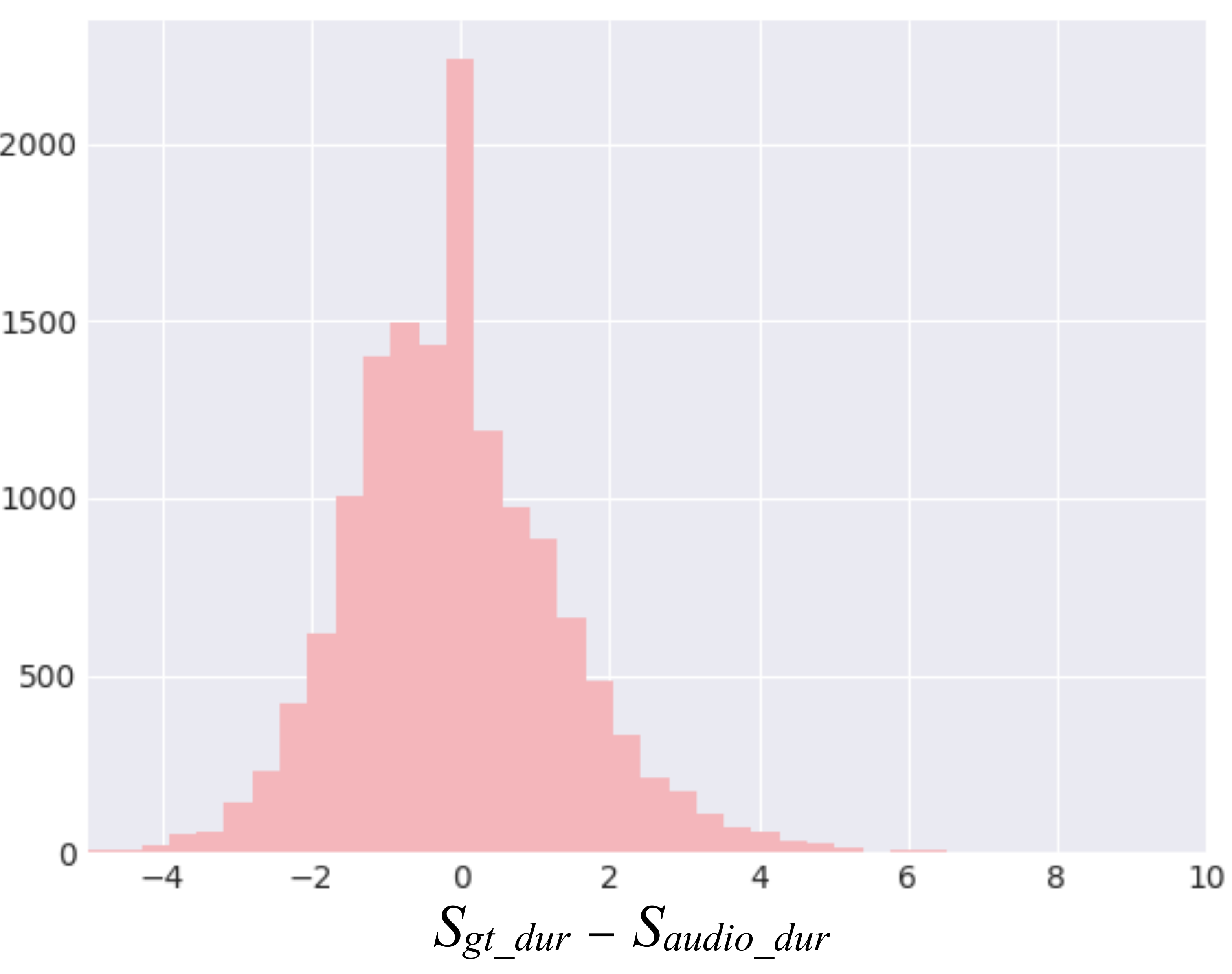}}
    \caption{\textbf{\boldmath S$_{gt}$ vs.~S$_{audio}$}: We plot the distribution of temporal shifts between 
        ground-truth (S$_{gt}$) and audio-aligned (S$_{audio}$) subtitles
        on the training split of the BSL-1K$_{aligned}$ dataset by showing the differences in subtitle
        (a) start times and (b) duration.
        We observe the difficulty of the subtitle alignment task: (i) there is no fixed shift between 
        ground-truth and audio-aligned subtitle timings, and (ii) the subtitle duration varies between spoken and signed languages.}
    \label{fig:shift}
\end{figure}

\begin{table}
    \centering
    \footnotesize
    \setlength{\tabcolsep}{2pt}
    \resizebox{0.99\linewidth}{!}{
    \begin{tabular}{lrrrrrc}
         \toprule
         & \#vids.
         & \#hours
         & \#subs
         & \#inst.
         & Vocab. 
         &  OOV\\
         \midrule
         Train & 20 &  14.4 & 13.8K & 128.1K & 8.6K & $\backslash$ \\
         Test (total) & 4 &  3.3& 2.0K & 18.6K & 2.8K & 726\\
         \hspace{3mm}signer$_{seen}$, genre$_{seen}$ & 1 & 0.7 &648 & 6.1K & 1.3K & 188\\
         \hspace{3mm}signer$_{seen}$, genre$_{unseen}$ & 1 & 0.9 &465 & 4.1K & 1.0K & 233 \\
         \hspace{3mm}signer$_{unseen}$, genre$_{seen}$ & 1  & 0.7 &506 & 5.6K & 1.1K & 99\\
         \hspace{3mm}signer$_{unseen}$, genre$_{unseen}$ & 1 & 1.0 & 360 & 2.8K & 882 & 234 \\
         \bottomrule
    \end{tabular}
    }
    \caption{\textbf{BSL-1K$_{aligned}$:} number of videos, hours, subtitles, word instances, vocabulary size and number of out-of-vocabulary (OOV) words. 
    }
    \label{tab:dataset}
\end{table}

\begin{table}
    \centering
    \footnotesize
    \resizebox{0.99\linewidth}{!}{
    \begin{tabular}{lrrrrrc}
         \toprule
         & \#vids.
         & \#hours
         & \#subs
         & \#inst.
         & Vocab. 
         &  OOV\\
         \midrule
         Train & 191 & 22.9 & 33.7K &261.5K & 7.5K & $\backslash$ \\
         Val & 15 & 1.5 & 2.6K & 18.1K & 1.8K & 196 \\
         Test & 21 & 2.6 & 3.8K & 27.3K & 2.4K & 369 \\
         \bottomrule
    \end{tabular}
    }
    \caption{\textbf{BSL Corpus:} number of videos,  hours, subtitles, word instances, vocabulary size and number of out-of-vocabulary (OOV) words in the dataset's splits. 
    }
    \label{tab:dataset-bslcorpus}
\end{table}

\noindent\textbf{BSL-1K$_{aligned}$} is a subset of BSL-1K~\cite{Albanie20}
which we manually annotated for subtitle alignment.
The subset contains 24 episodes %
covering a number of different television programmes (cooking, 
nature, travel and reality shows), corresponding to 17.7 hours of BSL content of 3 different signers with 16K %
subtitles. The subtitles were originally aligned to the audio, but
we have manually aligned them
to the signing.
The unaligned subtitles (i.e.\ those that are synchronised with the audio track, rather than the signing) differ from the signing-aligned subtitles in both start time and duration. In particular, Fig.~\ref{fig:shift}, shows that there is no fixed shift or temporal scaling that can be applied to transform audio-synchronised subtitles to their signing-aligned counterparts. We note that the differences exhibit an approximately Gaussian distribution, with the exception of an accentuated peak at 0 in Fig.~\ref{fig:distribution_dur}---if the duration of the subtitle is approximately correct, annotators tend not to further refine the boundaries.
The subtitles cover a total of 147K word instances for 
a vocabulary size of 9.4K in spoken English. We divide the data into 20 training episodes and 4 test episodes. The test episodes are chosen to evaluate the alignment model in different settings: seen/unseen signer and seen/unseen programme genre (which affects the number of out-of-vocabulary words) as 
shown in Tab.~\ref{tab:dataset}. The manual alignment of subtitles to signing content for the 24 episodes was performed over 
approximately 200 hours by
native BSL annotators
using the open-source VIA tool~\cite{Dutta19a}. %

\noindent\textbf{BSL Corpus}~\cite{schembri2013building,bslcorpus17} is a public dataset of videos of deaf signers gathered from several regions across the UK and accompanied by a variety of linguistic annotations. For our task, we employ the \textit{FreeTranslation} annotation tier, which provides written English subtitles to accompany portions of the \textit{Conversation} and \textit{Interview} subsets of the corpus. In total, the annotations cover a total of 227 videos after cropping to include a single signer. Of these, 141 are sourced from the \textit{Interview} subset and 86 videos are sourced from the \textit{Conversation} subset. For consistency with prior work, we follow the train, validation and test partition employed by~\cite{Renz21a,Albanie20}.
However, since this partition does not fully span the dataset, we add any dataset instances that were not present in the partition to the training set. Dataset statistics on the resulting train, validation and test partition, including the total number of hours, subtitles and vocabulary spanned by the data, are given in Tab.~\ref{tab:dataset-bslcorpus}. Unlike BSL-1K, the subtitles in this dataset are aligned to signing, and the translation direction is from sign language to English.
We therefore simulate unaligned data by perturbing the subtitle locations in our experiments.

\noindent\textbf{Evaluation metrics.}
We consider two main evaluation metrics: (i) frame-level accuracy, and (ii) $F1$-score.
For the $F1$-score, hits and misses of subtitle alignment to sign language video are counted under three temporal overlap thresholds
(IoU~$\in~\{0.1, 0.25, 0.50\}$) between predicted S$_{pred}$ and manually aligned S$_{gt}$ subtitles, denoted as F1@.10, F1@.25, F1@.50, respectively.

\subsection{Comparison to baselines} \label{subsec:baselines}

\noindent\textbf{Simple temporal shift baseline (}S$_{audio}^\text{+}$\textbf{).}
As a first baseline we use the shifted audio-aligned subtitles S$_{audio}^\text{+}$.

\noindent\textbf{Prosodic cues baseline (Bull et al.~\cite{bull2020}).} We compare to the state of the art on subtitle-unit segmentation, which is a model based on 2D body keypoints. In contrast to our framework, this method only uses visual prosodic cues and does not use semantic information from the query subtitle. It has been trained on a large-scale
sign language corpus with aligned subtitles, and the pretrained model is public.
The model consists of ST-GCN~\cite{stgcn} and BiLSTM layers and segments sign language video into subtitle units.
However, this is a different task than alignment, i.e.\ segments have no correspondence to subtitles.
To obtain an association from each predicted segment to a subtitle, we align
the shifted subtitles S$_{audio}^\text{+}$
to a subtitle-unit segmentation of \cite{bull2020} using DTW, where the cost of alignment is the temporal distance. %

\noindent\textbf{Heuristic baseline based on sparse sign spottings.}
Inspired by previous works that approached the alignment task through sparse 
correspondences~\cite{farhadi2006aligning}, we implement a heuristic approach to align the subtitles using a 
combination of sign spotting and active signer detection. Sign spotting, performed by \cite{Albanie20,Momeni20b},
searches in the temporal 
vicinity of each audio-synchronised subtitle (the search window is constructed by padding the original subtitle by 
four seconds
at each end) for individual sign instances corresponding to words that appear in the 
subtitle.
From these sparse sign localisations, we perform subtitle alignment in four stages.
First, we segment the episode into 
sequences that contain active signing, following~\cite{albanie21_seehear}.
Second, for any subtitle containing words that were spotted in the signing (assigned a posterior probability of 0.8 or 
greater by the model of~\cite{Momeni20b}), we shift the subtitle such that its centre falls on the mean position of the 
spotted signs. Third, we transform all subtitles without spottings by affine transformations such that they fall within 
the ``gaps'' between those subtitles that contained spotted signs, while preserving ordering (we use one such 
transformation per gap). Finally, we expand the duration of subtitles locally (applying a single scaling factor to each 
subtitle) in left to right ordering, such that they maximally fill the active signing segments predicted by the first stage. 

A comparison of our model to the above baselines is given in Tab.~\ref{tab:baseline}. The simple temporal shift baseline and 
the heuristic baseline based on sparse sign spottings perform similarly, but are a significant improvement over the non-shifted 
subtitles S$_{audio}$. Using prosodic cues through the model of~\cite{bull2020} results in a slight improvement over these two 
baselines. Our model significantly outperforms all baselines by exploiting the subtitle text to find the associated video 
segment. Indeed, when providing random subtitle text during training, our model fails to outperform baseline F1 scores.
Using DTW to resolve overlaps in predicted subtitles boosts our model performance.

A breakdown of our results by test episode is provided in Tab.~\ref{tab:breakdown_episodes}. Our model tends to result in 
larger improvements over the S$_{audio}^\text{+}$ baseline for signers seen in the training episodes, but still outperforms 
the S$_{audio}^\text{+}$ baseline for unseen signers in unseen genres. More training data would be needed to better 
generalise to unseen signers.

\begin{table}
    \centering
    \setlength{\tabcolsep}{6pt}
    \resizebox{0.99\linewidth}{!}{
    \begin{tabular}{lcccc}
        \toprule
        Method & frame-acc & F1@.10 & F1@.25 & F1@.50 \\
        \midrule
        S$_{audio}$ & 44.67 & 45.82 & 30.51 & 12.57 \\
        S$_{audio}^\text{+}$ & 60.76 & 71.69 & 60.74 & 36.10 \\
        \midrule
        Sign-spotting heuristics & 61.71 & 69.23 & 59.60 & 36.04 \\
        Bull et al.~\cite{bull2020} & 62.14 & 73.93 & 64.25 & 38.16 \\
        \midrule
        SAT (random subtitle) & 65.52 & 70.30 & 60.36 & 40.04 \\
        SAT w/out DTW & 65.81 & 74.32 & 64.69 & 41.27 \\
        SAT & \textbf{68.72} & \textbf{77.80} & \textbf{69.29} & \textbf{48.15} \\
        \bottomrule
    \end{tabular}
    }
    \caption{\textbf{Comparison to baselines:}
        We show significant improvements by
        training a Subtitle Aligner Transformer (SAT)
        over several baselines.
        Moreover, randomly shuffling subtitles obtains
        poor performance, demonstrating that our model does
        indeed rely on token embedding, and does not simply learn prosodic cues
        to align the subtitles.
        We obtain a further boost
        by correcting the overlaps of our predicted subtitles using DTW.
    }
    \label{tab:baseline}
\end{table}

\begin{table}
    \centering
    \setlength{\tabcolsep}{6pt}
    \resizebox{0.99\linewidth}{!}{
    \begin{tabular}{rrlcccc}
        \toprule
        \multicolumn{2}{c}{Test episode} \\
        signer & genre & Method & frame-acc & F1@.10 & F1@.25 & F1@.50  \\
        \midrule
        $seen$ & $seen$ & S$_{audio}^\text{+}$ & 45.48 & 66.92 & 55.02 & 31.84 \\
        & & SAT & \textbf{60.23} & \textbf{77.74} & \textbf{68.47} & \textbf{49.00} \\
        \midrule
        $seen$ & $unseen$ & S$_{audio}^\text{+}$ & 64.31 & 74.84 & 64.73 & 34.19 \\
        & & SAT & \textbf{72.56} & \textbf{81.29} & \textbf{74.19} & \textbf{52.47} \\
        \midrule
        $unseen$ & $seen$ & S$_{audio}^\text{+}$ & 56.30 & \textbf{80.79} & 69.70 & 44.95 \\
        & & SAT & \textbf{63.68} & 80.32 & \textbf{72.40} & \textbf{52.82} \\
        \midrule
        $unseen$ & $unseen$ & S$_{audio}^\text{+}$ & 71.84 & 63.29 & 53.16 & 33.76 \\
        & & SAT & \textbf{74.93} & \textbf{69.76} & \textbf{59.92} & \textbf{34.32} \\
        \bottomrule
    \end{tabular}
    }
    \caption{\textbf{Performance breakdown by test episode:} 
    Our model improves upon the S$_{audio}^{\text{+}}$ baseline for all the combinations of seen/unseen for signer and genre. The improvements however are greater in the test episodes where the signer has been seen during training.
    }
    \label{tab:breakdown_episodes}
\end{table}

\subsection{Ablation study} \label{subsec:ablations}
We ablate the effects of inputting the prior estimate S$_{prior} = S_{audio}^{\text{+}}$ to the model,
the size of the search window, modifying the text input to the encoder, pretraining on sign localisation and alternative model formulations. Some additional ablations are presented in Sec.~\ref{app:sec:experiments} of the appendix.

\noindent\textbf{Knowledge of S$_{prior}$.}
We experiment with several versions of
inputs as additional information to the alignment task.
Tab.~\ref{tab:ablations_reference} summarises the results.
We first observe a significant drop in performance
when S$_{prior}$ is not provided (48.15 vs 30.66 F1@.50),
suggesting that the position and duration of the corresponding
audio content allows an approximate localisation cue,
enabling the model to refine this via a series of attention layers.
Inputting the 3.2 seconds shifted subtitle timings (S$_{prior}=S_{audio}^{\text{+}}$) performs better than inputting the audio-aligned subtitle timings (S$_{prior}=S_{audio}$).
Moreover, we carry out two additional experiments
to investigate whether this cue provides a position
prior or a duration prior. First,
we always input the subtitle timing centred
with respect to the search window. The poor
performance of this model suggests the
importance of the position. Second,
we preserve the shifted location, but randomly
change the input subtitle duration at training time by up to 2s.
This slightly reduces the performance, therefore duration cues seem less essential for the model than location cues. 

\begin{table}
    \centering
    \setlength{\tabcolsep}{6pt}
    \resizebox{0.99\linewidth}{!}{
    \begin{tabular}{lcccc}
        \toprule
        Additional input & frame-acc & F1@.10 & F1@.25 & F1@.50 \\
        \midrule
        w/out S$_{audio}$ & 61.37 & 59.03 & 49.35 & 30.66 \\
        w/ S$_{audio}$ & 67.81 & 74.69 & 66.53 & 45.10 \\
        w/ S$_{audio}^{\text{+}}$ 3.2-sec shift & \textbf{68.72} & \textbf{77.80} & \textbf{69.29} & \textbf{48.15} \\
        \midrule
        w/ S$_{audio}$ centre position & 61.40 & 58.07 & 51.13 & 35.01 \\
        w/ S$_{audio}^{\text{+}}$ rand. duration & 68.61 & 75.10 & 66.84 & 46.72\\
        \bottomrule
    \end{tabular}
    }
    \caption{\textbf{Inputting S$_{prior}$ variants:}
        Without information on the approximate position and duration of the subtitle, our model fails to improve upon our baseline methods. In particular, when setting the input S$_{prior}$ to be systematically in the centre of the search window and with the duration of S$_{audio}$, model performance is poor. When using S$_{audio}^{\text{+}}$ in its correct location in the search window, but varying the duration randomly of up to 2s, performance is relatively high. This suggests the position is a stronger cue than duration.
    }
    \label{tab:ablations_reference}
\end{table}

\noindent\textbf{Size of the search window $T$.}
In Tab.~\ref{tab:ablations_window},
we report the performance against different choices for input duration $T$
We conclude that larger search windows generally
improve performance, at the cost of computational complexity.
This might be due to increased supervision, since with larger windows the training
sees more negative examples, as well as due to better
coverage at test time. A too short window size inhibits recovery of the correct location, if the correct location falls outside of the window boundaries.
In all our experiments, we use 20-second windows. 

\begin{table}
    \centering
    \setlength{\tabcolsep}{6pt}
    \resizebox{0.99\linewidth}{!}{
    \begin{tabular}{lcccc}
        \toprule
        Window size & frame-acc & F1@.10 & F1@.25 & F1@.50 \\
        \midrule
        8 sec & 66.98 & 73.12 & 64.66 & 44.13 \\
        12 sec & 68.63 & 75.52 & 67.56 & 47.29 \\
        16 sec & 68.51 & 76.18 & 68.63 & 48.10 \\
        20 sec & \textbf{68.72} & \textbf{77.80} & \textbf{69.29} & \textbf{48.15} \\
        \bottomrule
    \end{tabular}
    }
    \caption{\textbf{Search window size $T$:}
        We vary $T$ between 50 and 125 frames (corresponding to 8- and 20-second inputs, respectively). Larger windows tend to perform better, possibly due to 
        increased contextual information and the fact that the difference between S$_{audio}$ and the aligned subtitle 
        S$_{gt}$ can be in the order of 10s.
    }
    \label{tab:ablations_window}
\end{table}

\noindent\textbf{Effect of text input to the encoder.} 
 We perform a series of ablations regarding the text encoding, including: no text augmentations, adding extra positional encodings to the BERT text features (as described in appendix Sec.~\ref{app:sec:implementation}), and using the sentence embedding only (the output embedding corresponding to the BERT ``CLS" token) instead of the sequence of individual token embeddings. Tab.~\ref{tab:augment} presents the results on BSL-1K$_{aligned}$
with these text ablations. Augmenting the subtitle text improves performance, while adding extra positional encodings or using the sentence embedding degrades performance. 

\begin{table}
    \centering
    \setlength{\tabcolsep}{6pt}
    \resizebox{0.99\linewidth}{!}{
    \begin{tabular}{lcccc}
        \toprule
        Method & frame-acc & F1@.10 & F1@.25 & F1@.50 \\
        \midrule
        w/o augmentations & 67.35 & 75.72 & 66.85 & 45.31 \\
        w/ augmentations & \textbf{68.72} & \textbf{77.80} & \textbf{69.29} & \textbf{48.15} \\
        w/ aug. + positional enc. &  68.21 & 74.89 & 67.14 & 46.36 \\
        w/ aug. sentence emb. & 66.18 & 72.99 & 63.71 & 41.71 \\
        \bottomrule
    \end{tabular}
    }
    \caption{\textbf{Text ablations:}
    As a data augmentation step during training, we shuffle the words in 50\% of the subtitles and add or delete up to 2 words in the subtitle. This results in a large performance gain. Adding positional encodings to the BERT text features does not improve our model. Using sentence embeddings instead of token embeddings for the subtitle query degrades performance. 
    }
    \label{tab:augment}
\end{table}

\noindent\textbf{Effect of sign localisation pretraining.}
As explained in Sec.~\ref{subsec:wordpretraining},
we initially pretrain our model for temporal localisation of individual signs.
In Tab.~\ref{tab:pretraining}, we measure
the effect of this pretraining on a large set of word-video training pairs,
and conclude that it provides a good initialisation
for finetuning on long subtitles.

\begin{table}
    \centering
    \setlength{\tabcolsep}{6pt}
    \resizebox{0.99\linewidth}{!}{
    \begin{tabular}{lcccc}
        \toprule
        Pretraining & frame-acc & F1@.10 & F1@.25 & F1@.50 \\
        \midrule
        w/o word pretraining & 67.26 & 76.18 & 66.19 & 42.47 \\
        w/ word pretraining & \textbf{68.72} & \textbf{77.80} & \textbf{69.29} & \textbf{48.15} \\
        \bottomrule
    \end{tabular}
    }
    \caption{\textbf{Pretraining for sign localisation:}
    By pretraining our model to locate individual words %
    within a given temporal window, we boost performance of subtitle alignment.}
    \label{tab:pretraining}
\end{table}

\noindent\textbf{Model formulation.} We consider an alternative version of the Transformer model, inspired by the DETR model in~\cite{carion2020end} for object detection in images.
This model inputs image features into the Transformer encoder and text query into the Transformer decoder.
Similarly, we input the sign language video features into the Transformer encoder. On the decoder side, we input the subtitle 
text features as well as either (i) the start and end times or (ii) the shift and scale of the shifted subtitles 
S$_{audio}^{\text{+}}$ relative to the temporal window. We then consider the problem of subtitle alignment as a regression 
problem, and aim to predict (i) the start and end times or (ii) the shift and scale of the subtitle relative to the temporal window.
As a further ablation, we also consider the same model architecture (with subtitle features and the start and end times as decoder input), but outputting a fixed binary vector of length $T$, which we train with a binary classification objective (as in SAT).

The results in Tab.~\ref{tab:loss} suggest
that our proposed approach with video features as input to the Transformer decoder enables significantly better learning, 
perhaps by providing a one-to-one mapping between video
inputs and the frame-wise outputs. 
Another possible explanation for our proposed model's superiority is that it outputs alignment scores between subtitles and individual frames which allows for better conflict resolution strategies for overlapping subtitle predictions.

\begin{table}
    \centering
    \setlength{\tabcolsep}{6pt}
    \resizebox{0.99\linewidth}{!}{
    \begin{tabular}{llcccc}
        \toprule
        Prior input & Loss & frame-acc & F1@.10 & F1@.25 & F1@.50 \\
        \midrule
        shift/scale & shift/scale regress. & 59.23 & 70.55 & 59.00 & 33.71 \\
        start/end & start/end regress. & 60.04 & 72.20 & 60.41 & 34.33 \\
        start/end & binary classif. & 60.48 & 74.05 & 62.75 & 35.07\\
        binary & binary classif. (SAT) & \textbf{68.72} & \textbf{77.80} & \textbf{69.29} & \textbf{48.15} \\
        \bottomrule
    \end{tabular}
    }
    \caption{\textbf{Model formulation:}
        We present an ablation where we experiment with a DETR-style Transformer model~\cite{carion2020end}. Video features are inputs to the Transformer encoder, and the subtitle query is fed to the Transformer decoder. Moreover, on the decoder side, we input either the start and end times or the shift and scale of the shifted subtitles S$_{audio}^{\text{+}}$ relative to the temporal window, and use a regression model to predict the true values. This model fails to produce satisfactory results. Changing the regression model to a classification one by instead predicting a binary vector of length $T$ (as in the SAT model) results in a small improvement; however SAT outperforms all the alternative models with a large margin.
    }
    \label{tab:loss}
\end{table}

\subsection{Performance on a different dataset}
\label{subsec:bslcp}

\begin{figure*}[t]
    \centering
      \includegraphics[width=\textwidth]{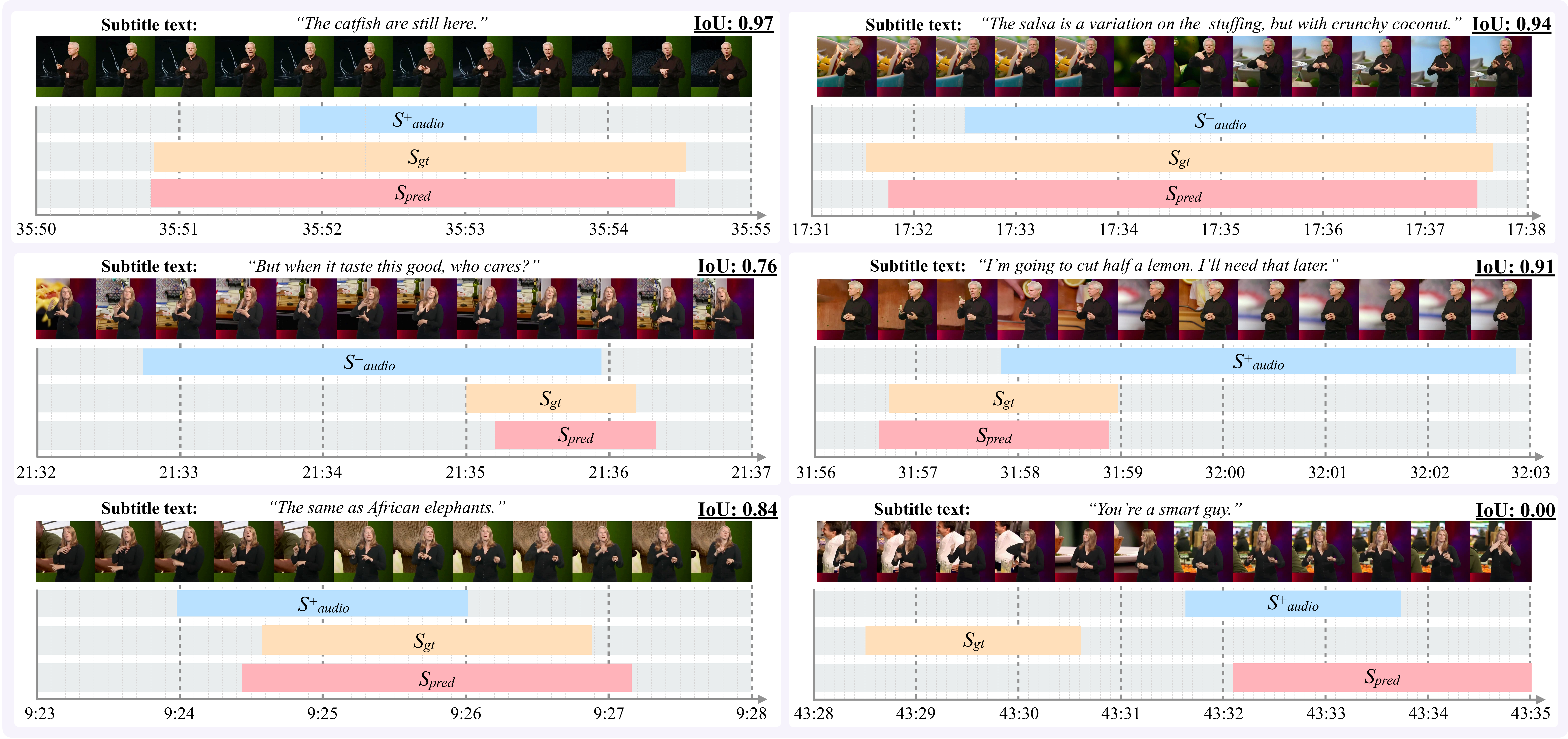}
    \caption{\textbf{Qualitative results:} This figure shows short time windows of 5s (left) or 7s (right) with shifted audio-aligned subtitles (S$_{audio}^{\text{+}}$), ground truth signing-aligned subtitles (S$_{gt}$) and our predicted signing-aligned subtitles (S$_{pred}$). In practice, we input 20 seconds of video during training and testing as our search window.
    }
    \vspace{-0.4cm}
    \label{fig:qualitative}
\end{figure*}

We demonstrate our model's performance on the BSL Corpus~\cite{schembri2013building,bslcorpus17}.
The subtitles in this dataset are aligned to the sign language, and so we randomly shift and scale the subtitles in order to create artificial training data. We then train our SAT model to learn the correct alignment of subtitles to video in the BSL Corpus. We train the model (i)~without any pretraining, (ii)~with only word pretraining (on BSL-1K) and (iii)~with SAT pretraining on BSL-1K$_{aligned}$. We report results in Tab.~\ref{tab:bslcorpus_results}.

At each subtitle, we apply a random shift following a normal distribution with standard deviation $\sigma_{\text{pos}}$ and a random change of duration of the subtitle also following a normal distribution with standard deviation $\sigma_{\text{dur}}$. Tab.~\ref{tab:bslcorpus_results} shows that our model is able to partially recover the correct original alignment. Larger shifts make it more difficult for our model to recover the correct original alignment, but random changes in subtitle duration seems to have less effect. This is consistent with the results in Tab.~\ref{tab:ablations_reference}, where changing the duration of S$_{audio}^{\text{+}}$ does not greatly impact results. Word pretraining on BSL-1K helps the model, but SAT pretraining on BSL-1K$_{aligned}$ does not. Word pretraining may help the SAT model recognise certain signs in BSL, but domain difference between BSL Corpus and BSL-1K$_{aligned}$ subtitles may explain why SAT pretraining on BSL-1K$_{aligned}$ does not lead to any significant gains on BSL Corpus.

\begin{table}
    \centering
    \setlength{\tabcolsep}{4pt}
    \resizebox{0.99\linewidth}{!}{
    \begin{tabular}{llcccc}
        \toprule
        \multicolumn{1}{l}{Rand. perturb.} \\
        ($\sigma_{\text{pos}}$, $\sigma_{\text{dur}}$) & Method & frame-acc & F1@.10 & F1@.25 & F1@.50  \\
        \midrule
        (3.5s, 1.5s) & Rand. shift \& scale & 63.24 & 37.13 & 26.54 & 12.47 \\
         & SAT w/out pretrain. & 73.73 & 51.51 & 43.33 & 27.98 \\
         & SAT pretrain. & {75.77} & {55.55} & {47.45} & {32.57} \\
         & SAT w/ word pretrain. & \textbf{76.29} & \textbf{57.65} & \textbf{50.35} & \textbf{34.54} \\
        \midrule
        (4.5s, 1.5s) & Rand. shift \& scale & 60.18 & 29.52 & 20.61 & 10.00 \\
        & SAT pretrain. & 73.69 & 48.41 & 41.34 & 28.06 \\
         & SAT w/ word pretrain. & \textbf{74.29} & \textbf{51.33} & \textbf{44.37} & \textbf{30.13}  \\
        \midrule
        (3.5s, 2s) & Rand. shift \& scale & 62.62 & 37.47 & 26.82 & 11.87 \\
         & SAT pretrain. & 75.79 & 55.31 & 47.24 & 32.89 \\
         & SAT w/ word pretrain. & \textbf{76.00} & \textbf{57.86} & \textbf{50.43} &
         \textbf{33.79}  \\
        \bottomrule
    \end{tabular}
    }
    \caption{\textbf{BSL Corpus:} 
    We show results on another dataset \cite{schembri2013building,bslcorpus17}
    with subtitles aligned to signing.
    We randomly shift and scale the correctly aligned subtitles in BSL Corpus
    to simulate unaligned data
    and then use our SAT model to recover the original correct alignments. Position is randomly shifted following a normal distribution with standard deviation $\sigma_{\text{pos}}$ and duration is randomly changed according to a normal distribution with standard deviation $\sigma_{\text{dur}}$.
    Our model is capable of learning to align subtitles on this data. Word pretraining on BSL-1K increases performance, but pretraining the SAT model on BSL-1K$_{aligned}$ (SAT~pretrain.) does not result in further gains. 
    }
    \label{tab:bslcorpus_results}
\end{table}

\subsection{Qualitative analysis}
\label{subsec:qualitative}

Fig.~\ref{fig:qualitative} illustrates
several test examples on BSL-1K$_{aligned}$. The timeline shows the ground truth
alignment (S$_{gt}$), our prediction (S$_{pred}$), as well as the S$_{audio}^{\text{+}}$
baseline, alongside a sample of video frames and the query subtitle text. While the shifted baseline
S$_{audio}^{\text{+}}$ provides an approximate position,
it is largely unaligned. Our model effectively
learns to attend to both visual and textual cues.
A typical failure mode happens when the prior position encoding
is significantly far from the ground truth (see Fig.~\ref{fig:qualitative}
bottom right). For additional qualitative examples on BSL Corpus, we refer to Fig.~\ref{app:fig:bslcp}
of the appendix.

\section{Conclusion}
\label{sec:conclusions}

We presented a Transformer-based
approach to synchronise subtitles
with sign language video content
in interpreted data.
We showed that knowledge
of subtitle content is essential
to effectively align subtitles to signing.
We hope that our work will be
a stepping stone to obtain video-subtitle
pairs that allow training of unconstrained
machine translation systems for sign languages.
Furthermore, our approach is potentially
applicable to other domains, such as temporal grounding of
sentences. We refer to Sec.~\ref{app:sec:impact} of the appendix
for a discussion
on the broader impact on the community. 

\noindent\textbf{Acknowledgements.}
This work was supported by EPSRC grant ExTol and a Royal Society Research Professorship.
We thank Tom Monnier, %
Himel Chowdhury, Abhishek Dutta, Ashish Thandavan, 
Annelies Braffort, %
Michèle Gouiffès %
and Igor Garbuz  %
for their help.

{\small
\bibliographystyle{ieee_fullname}
\bibliography{shortstrings,vgg_local,references}
}

\clearpage
\section*{Appendix}
\renewcommand{\thefigure}{A.\arabic{figure}} %
\setcounter{figure}{0} 
\renewcommand{\thetable}{A.\arabic{table}}
\setcounter{table}{0}

\appendix

\vspace{10pt}

We provide
further implementation details (Sec.~\ref{app:sec:implementation}),
additional qualitative results (Sec.~\ref{app:sec:qualitative}),
additional experiments (Sec.~\ref{app:sec:experiments}),
and a broader impact statement (Sec.~\ref{app:sec:impact}).

\section{Implementation details}
\label{app:sec:implementation}

\noindent\textbf{Text embeddings.}
For the text embeddings, we use a pretrained BERT model from HuggingFace\footnote{\url{https://huggingface.co/bert-base-uncased}} with a standard architecture of 12-layers, 12-heads and 768 model size. The model is pretrained on BookCorpus\footnote{\url{https://yknzhu.wixsite.com/mbweb}} and English Wikipedia\footnote{\url{https://en.wikipedia.org}}. 

\noindent\textbf{Positional encodings.}
For the input to the video encoder, we use $512$-dimensional sinusoidal positional encodings as in \cite{vaswani2017attention}. The positional encodings are added to the video features before feeding to the Transformer.

\noindent\textbf{Output thresholding.}
The output of our model is a temporal sequence of predictions between 0 and 1. For the single-subtitle SAT model, we 
consider the start of the subtitle to be the first time when the prediction is above $\tau=0.5$ and the end of the subtitle to be the last time when the prediction is above $\tau=0.5$ in the search window. When we apply a global alignment step with DTW to 
correct for overlapping subtitles, we no longer use these thresholds, but rather the temporal sequence of predictions between 
0 and 1 using the method described in the main paper.

\noindent\textbf{Training details.}
We use the Adam optimiser with a batch size of 64.
We train with a learning rate of $10^{-5}$ at the word-pretraining stage,
and of $5\times10^{-6}$ at finetuning with subtitles. 
At the word pretraining stage, the model is trained over 5 epochs. 
In one epoch of word pretraining, there are approximately 700K sign instances (including sign spotting both with mouthings and dictionaries). 
At this point the word alignment model obtains a frame-level accuracy of 30.38\% and F1@50 of 40.75\% on the 1630 sign instances of the test set episodes. 
During full-sentence finetuning, the model is trained over 80 epochs.

\begin{figure*}
    \centering
      \includegraphics[width=0.96\textwidth]{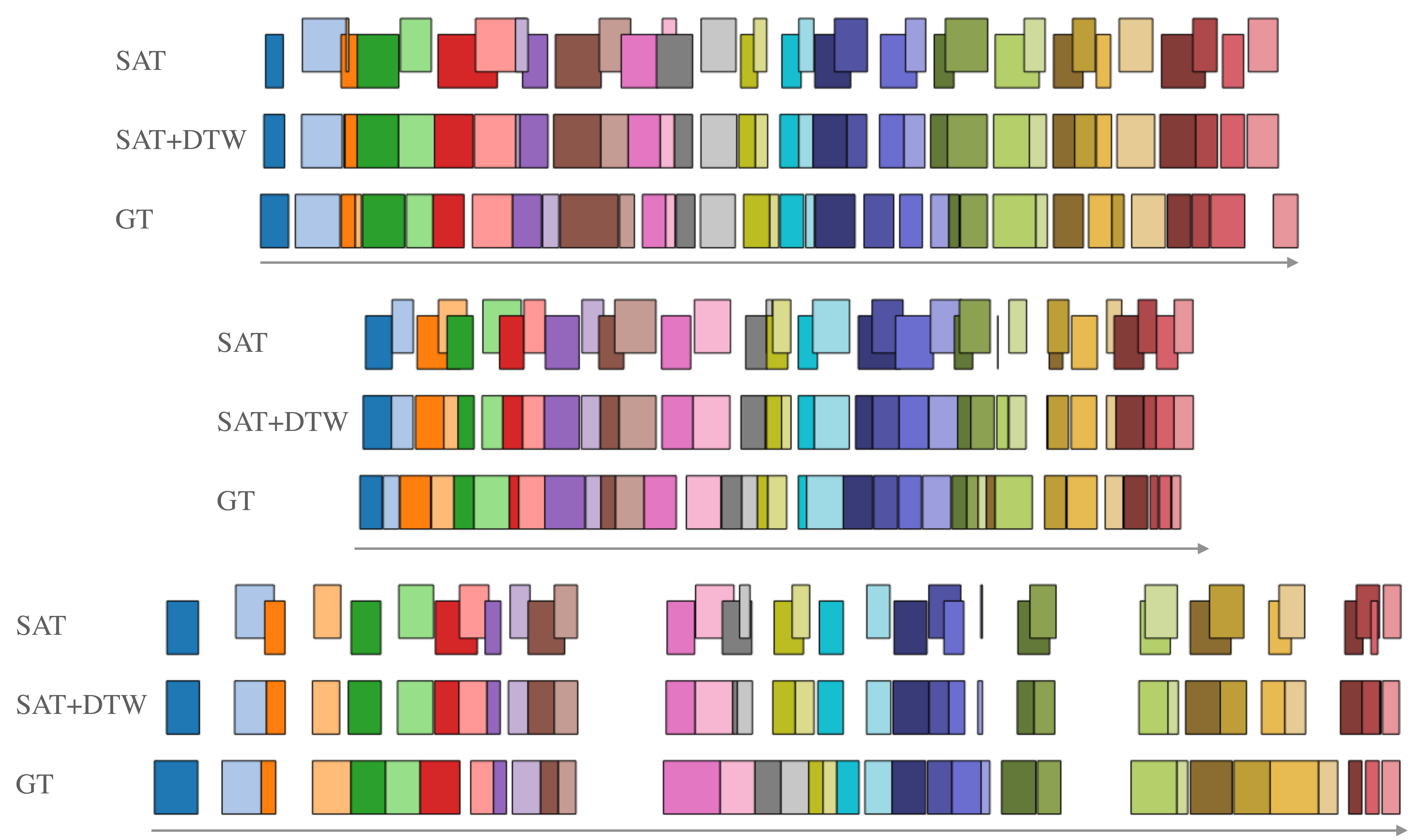}
    \caption{\textbf{DTW:} Our SAT model predicts the locations of subtitles independently of each other, and thus there can be overlaps in subtitle localisations. Using a global alignment step with DTW, we resolve these overlaps and improve performance. %
    }
    \label{app:fig:dtw}
\end{figure*}

\section{Additional qualitative analysis}
\label{app:sec:qualitative}

\begin{figure}
\vspace{-15pt}
    \centering
    \includegraphics[width=0.99\linewidth]{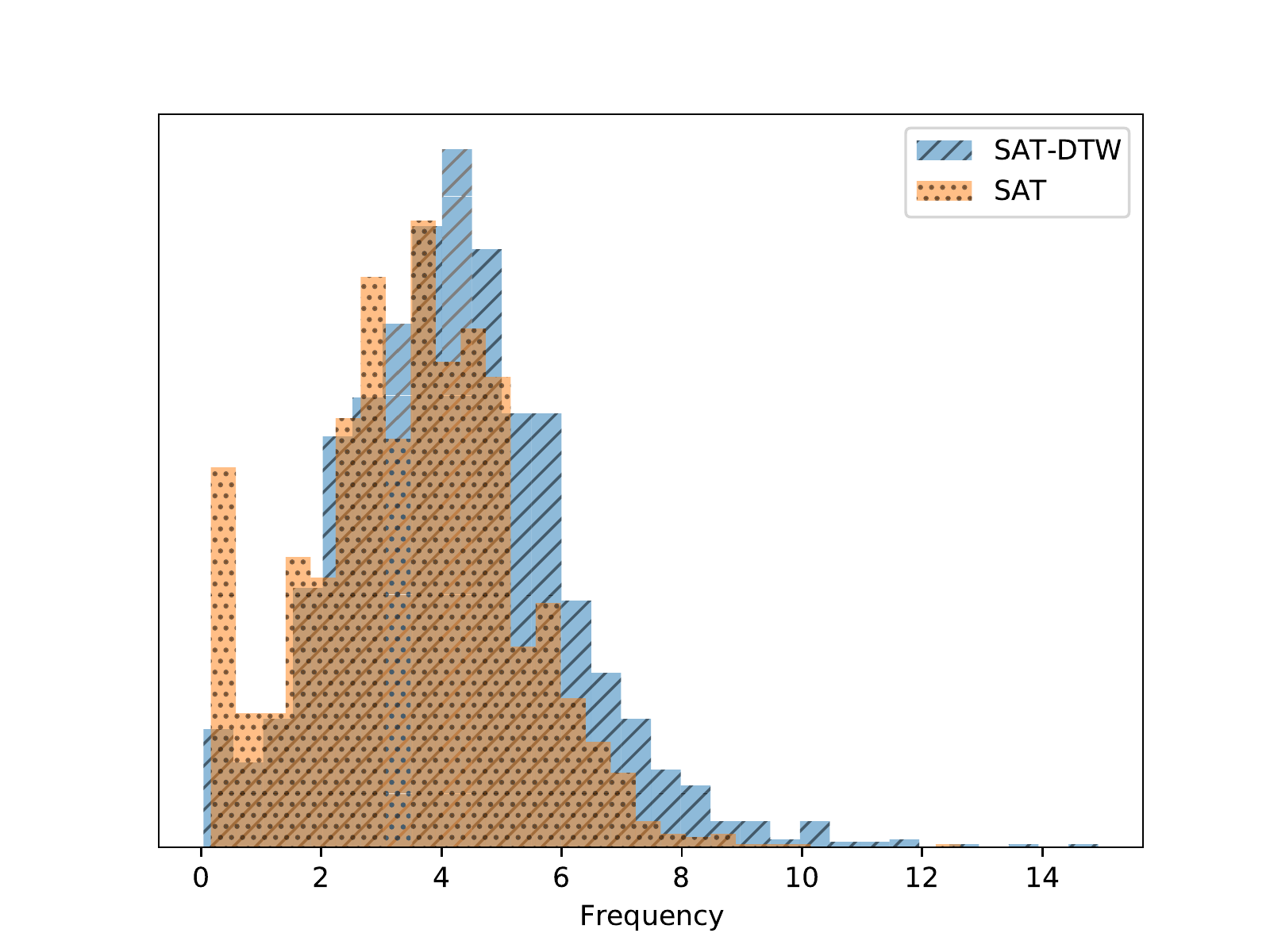}
    \caption{\textbf{Duration before and after DTW}: The median duration of S$_{gt}$ is 3.3s. Before DTW, the median duration of our predicted subtitles is 4.1s, but after DTW the median duration is reduced back down to 3.5s by resolving conflicts in overlapping subtitles.}
    \label{app:fig:histogram_duration}
\end{figure}

\noindent\textbf{Effect of global alignment with DTW.}
In Fig.~\ref{app:fig:dtw}, we present results before and after
the global alignment with DTW on a long timeline. We observe
that the single-subtitle Transformer model produces
overlapping regions between consecutive subtitles
which are resolved after the global DTW stage.
Consequently, we see that the overall duration of subtitles decreases
after DTW (see Fig.~\ref{app:fig:histogram_duration}). During the DTW stage, we order subtitles by their predicted order, not by the original order of S$_{audio}$. Indeed, in BSL-1K$_{aligned}$, 1.6\% of subtitles in S$_{gt}$ do not respect the original order of S$_{audio}$. On the test set, 1.6\% of subtitles in S$_{pred}$ switch position with respect to S$_{audio}$. 

\begin{figure*}
    \centering
     \includegraphics[width=\textwidth]{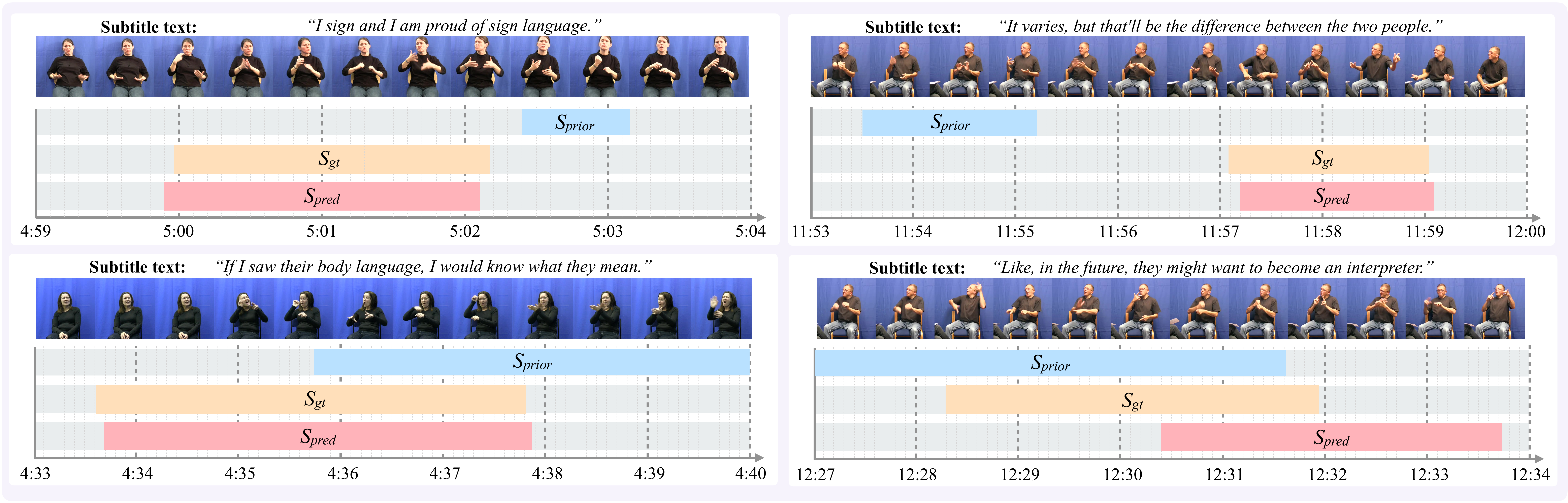}
    \caption{\textbf{Qualitative results on BSL Corpus:} This figure shows short time windows of 5s and 7s with shifted and rescaled subtitles (S$_{prior}$), ground truth aligned subtitles (S$_{gt}$) and our predicted subtitles (S$_{pred}$). In practice, we input 20 seconds of video during training and testing for our search window. The shifted and rescaled subtitles (S$_{prior}$) are created using a random shift with standard deviation of 3.5s and a random change in length of standard deviation 1.5s.
    }
    \label{app:fig:bslcp}
\end{figure*}

\noindent\textbf{Results on BSL Corpus.}
Fig.~\ref{app:fig:bslcp} demonstrates
qualitative results on BSL Corpus.

\section{Additional experiments}
\label{app:sec:experiments}

We perform ablations to evaluate the influence
of our data augmentations and the encoding
choice for the subtitle text.

\noindent\textbf{Text encoding choice.}
We experiment with word2vec~\cite{mikolov2013distributed} encodings
for subtitle words instead of BERT as used
in the main paper experiments. We use the pretrained word2vec model from~\cite{miech2020end}, forming sentence embeddings by max pooling the encodings of all words over the channel dimension.
In Tab.~\ref{app:tab:wordemb}, we see that this results in lower performance compared to using the BERT encodings. We 
hypothesize that this is due to word2vec using a limited vocabulary, ignoring word order, and lacking the large-scale 
pretraining of the BERT model.

\begin{table}[h]
    \centering
    \setlength{\tabcolsep}{6pt}
    \resizebox{0.89\linewidth}{!}{
    \begin{tabular}{lcccc}
        \toprule
        Method & frame-acc & F1@.10 & F1@.25 & F1@.50 \\
        \midrule
        word2vec & 67.16 & 74.59 & 64.96 & 42.06  \\
        BERT & \textbf{68.72} & \textbf{77.80} & \textbf{69.29} & \textbf{48.15} \\
        \bottomrule
    \end{tabular}
    }
    \caption{\textbf{Text encoding:}
    We experiment with word2vec encodings instead of BERT
    to embed words in the subtitle.
    }
    \label{app:tab:wordemb}
\end{table}

\noindent\textbf{Amount of training data.}
By increasing the amount of training data, we improve performance of our model on the test set. Tab.~\ref{app:tab:amount_data} shows our results when training on random subsets of 25\%, 50\% and 75\% of the videos in our training data. For subset selection, we randomly sample 4 times, and report the average performance across 4 trainings, as well as the standard deviation.

\begin{table}[h]
    \centering
    \setlength{\tabcolsep}{6pt}
    \resizebox{0.99\linewidth}{!}{
    \begin{tabular}{lllll}
        \toprule
        \#training videos & frame-acc & F1@.10 & F1@.25 & F1@.50 \\
        \midrule
        5 &  66.62$^{\pm0.16}$ & 75.55$^{\pm0.86}$ & 66.04$^{\pm1.09}$ & 43.24$^{\pm0.81}$ \\
        10 & 67.40$^{\pm0.28}$ & 75.74$^{\pm0.25}$ & 66.60$^{\pm0.25}$ & 45.41$^{\pm0.88}$ \\
        15 & 67.71$^{\pm0.23}$ & 75.24$^{\pm0.43}$ & 66.29$^{\pm0.84}$ & 46.16$^{\pm0.66}$ \\
        20 & \textbf{68.72} & \textbf{77.80} & \textbf{69.29} & \textbf{48.15} \\

        \bottomrule
    \end{tabular}
    }
    \caption{\textbf{Amount of training data:}
    We train with a subset of our videos, using 5, 10, or 15 episodes
    instead of the total 20 used in the paper. We observe increased performance as we increase the training size.
    }
    \label{app:tab:amount_data}
\end{table}

\noindent\textbf{Sensitivity analysis.} During inference, we predict the location of a subtitle within a 20 second search window surrounding the location of S$_{audio}^{\text{+}}$. In order to analyse the sensitivity of the choice of search window, we shift the window by 1s, 3s and 5s at inference time. Tab.~\ref{app:tab:change_window} shows that the choice of window within a margin of a few seconds does not have a large impact on the results. 

However, if we keep the position of the search window constant and change the position of the prior estimate S$_{audio}^{\text{+}}$, then this has a significant effect on results. 
Tab.~\ref{app:tab:change_ref} shows the results of an experiment where we shift the prior estimate S$_{audio}^{\text{+}}$ by 1s, 3s and 5s at inference time.
The performance degrades when the model is given a worse prior as input, i.e., shifting S$_{audio}^{\text{+}}$.

\begin{table}
    \centering
    \setlength{\tabcolsep}{6pt}
    \resizebox{0.99\linewidth}{!}{
    \begin{tabular}{lcccc}
        \toprule
        Shift window & frame-acc & F1@.10 & F1@.25 & F1@.50 \\
        \midrule
        0s & \textbf{68.72} & \textbf{77.80} & \textbf{69.29} & 48.15 \\
        1s & 68.53 & 76.99 & 69.23 & 47.69 \\
        3s & 68.53 & 76.99 & 68.32 & 47.90 \\
        5s & 68.32 & 76.58 & 68.42 & \textbf{48.50} \\
        \bottomrule
    \end{tabular}
    }
    \caption{\textbf{Shifting search window:} We shift the search window at inference time by 1s, 3s and 5s. This does not have a major impact on results. 
    }
    \label{app:tab:change_window}
\end{table}

\begin{table}
    \centering
    \setlength{\tabcolsep}{6pt}
    \resizebox{0.99\linewidth}{!}{
    \begin{tabular}{lcccc}
        \toprule
        Shift prior & frame-acc & F1@.10 & F1@.25 & F1@.50 \\
        \midrule
        0s & \textbf{68.72} & \textbf{77.80} & \textbf{69.29} & \textbf{48.15} \\
        1s & 68.26 & 75.77 & 67.36 & 45.67  \\
        3s & 58.69 & 58.08 & 47.80 & 28.18  \\
        5s & 46.11 & 35.49 & 26.21 & 12.52 \\
        \bottomrule
    \end{tabular}
    }
    \caption{\textbf{Shifting prior estimate S$_{audio}^{\text{+}}$:} By shifting the location of the prior S$_{audio}^{\text{+}}$ at inference time by respectively 1s, 3s and 5s, the performance degrades.
    }
    \label{app:tab:change_ref}
\end{table}

\section{Broader impact}
\label{app:sec:impact}

The World Federation of the Deaf states that there are 70 million Deaf individuals world-wide using more than 200 sign languages.\footnote{\url{http://wfdeaf.org/our-work/}} Unfortunately, many technologies for spoken and written languages do not exist for signed languages. We hope that our work contributes towards addressing this imbalance by providing inclusive technologies for signed languages for several applications, discussed next.

One direct application of our method is an assistive subtitling tool for signing vloggers to align their subtitles (this technology is currently only available for spoken and written languages). A second application is to create bilingual written-signed corpora aligned at a sentence or phrase-like level. Such corpora can be used in contextual or concordance dictionaries, useful for translation or for language learning~\cite{kaczmarek2020use}. Moreover, they can be used as training data for translation between signing and written text. 
For context, note that machine translation---which can now be performed to an acceptable level in many written languages to enable cross-lingual access to content---remains far from human performance for sign languages~\cite{koller2020quantitative}. To enable progress for this task (and others that have been highlighted as important by members of Deaf communities), a key stumbling block is the availability of larger annotated datasets~\cite{bragg2019sign}. Our work aims to take steps towards addressing this challenge, since automatic subtitle alignment represents an important pre-processing step that has been performed manually for existing translation datasets, e.g.~\cite{Camgoz_2018_CVPR}.  However, scaling manual annotation to larger datasets is prohibitively expensive (as noted in the submission, aligning one hour of video takes approximately 10-15 hours of annotation time).

We note that there are also potential risks associated with our contributions. First, there is a chance with any computational advances in sign language modelling that it leads to increased surveillance of Deaf communities (and of content moderation more generally). Second, we note that our training data, obtained from public broadcast footage, may not be demographically representative of the population as a whole, and therefore is susceptible to bias. Additionally, the videos contain BSL interpreted from English, not original BSL content. Subtitle alignment may work less effectively for individuals who are not well-represented in the training data.

\end{document}